\documentclass[10pt,twocolumn,letterpaper]{article}

\usepackage{amsmath,amsfonts,bm}

\def\eqref#1{equation~\ref{#1}}

\def\1{\bm{1}}

\def\vc{{\bm{c}}}

\def\vx{{\bm{x}}}

\def\vz{{\bm{z}}}

\def\mA{{\bm{A}}}

\def\mC{{\bm{C}}}
\def\mD{{\bm{D}}}

\def\mZ{{\bm{Z}}}

\DeclareMathAlphabet{\mathsfit}{\encodingdefault}{\sfdefault}{m}{sl}
\SetMathAlphabet{\mathsfit}{bold}{\encodingdefault}{\sfdefault}{bx}{n}

\DeclareMathOperator*{\argmax}{arg\,max}
\DeclareMathOperator*{\argmin}{arg\,min}

\usepackage[pagenumbers]{cvpr} 

\definecolor{cvprblue}{rgb}{0.21,0.49,0.74}
\usepackage[pagebackref,breaklinks,colorlinks,allcolors=cvprblue]{hyperref}

\usepackage{cancel}
\usepackage{bigstrut}
\usepackage{multirow}
\usepackage{amsmath}
\usepackage{bbding}
\usepackage{titletoc}
\usepackage{algorithmic}
\usepackage{algorithm}

\newcommand{\trc}[1]{\textcolor[RGB]{227,23,13}{\textbf{#1}}}

\newcommand{\tgc}[1]{\textcolor[RGB]{50,205,50}{\textbf{#1}}}

\def\paperID{1610} 
\def\confName{CVPR}
\def\confYear{2025}

\title{Preventing Local Pitfalls in Vector Quantization via Optimal Transport}

\author{Borui Zhang\ , Wenzhao Zheng\ , Jie Zhou\ , Jiwen Lu\thanks{Corresponding author.}\\
{Department of Automation, Tsinghua University, China}\\
{\tt\small zhang-br21@mails.tsinghua.edu.cn; wenzhao.zheng@outlook.com; \{jzhou, lujiwen\}@tsinghua.edu.cn}
}

\begin{document}
\maketitle
\begin{abstract}
    Vector-quantized networks (VQNs) have exhibited remarkable performance across various tasks, yet they are prone to training instability,
    which complicates the training process due to the necessity for techniques such as subtle initialization and model distillation.
    In this study, we identify the local minima issue as the primary cause of this instability.
    To address this, we integrate an optimal transport method in place of the nearest neighbor search to achieve a more globally informed assignment.
    We introduce \textbf{OptVQ}, a novel vector quantization method that employs the Sinkhorn algorithm to optimize the optimal transport problem, thereby enhancing the stability and efficiency of the training process.
    To mitigate the influence of diverse data distributions on the Sinkhorn algorithm, we implement a straightforward yet effective normalization strategy.
    Our comprehensive experiments on image reconstruction tasks demonstrate that OptVQ achieves 100\% codebook utilization and surpasses current state-of-the-art VQNs in reconstruction quality.
    \footnote{Code: \url{https://github.com/zbr17/OptVQ}}
\end{abstract}

\vspace{-4mm}
\section{Introduction}

Vector quantization (VQ) is a widely utilized discretization technique that transforms data from continuous spaces into discrete tokens.
Building upon this, auto-encoders with vector quantization~\cite{van2017neural,esser2021taming} are designed to extract discrete representations from continuous spaces while maintaining high compressibility.
These models are characterized by an encoder-decoder architecture complemented by a codebook.
The encoder's role is to map input data into a continuous latent space, whereas the codebook functions as a reference for the conversion of data from continuous to discrete tokens.
As we enter the 2020s, the proliferation of large-scale models~\cite{brown2020language,achiam2023gpt} has catalyzed the exploration of universal approaches for modeling the distribution of diverse data modalities.
The paradigm of ``predicting the next token'' has emerged as a versatile modeling strategy.
In this context, the vector quantization technique plays a crucial role~\cite{esser2021taming,chang2022maskgit,baobeit2022,bai2024sequential} in bridging continuous-space data with discrete tokens. 
This enables large sequential models to process a variety of data modalities more effectively.

\begin{figure}[tbp]
    \centering
    \subfloat[\label{fig:head_a}]{
        \includegraphics[width=0.48\linewidth]{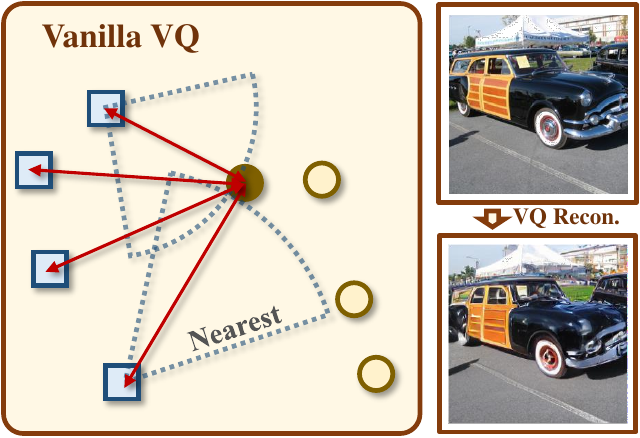}
    }
    \subfloat[\label{fig:head_b}]{
        \includegraphics[width=0.48\linewidth]{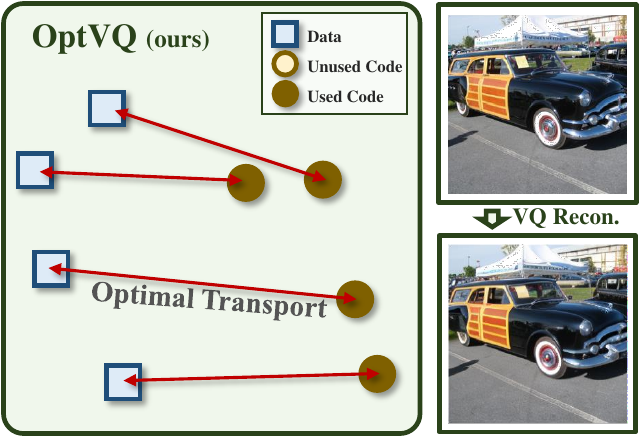}
    }
    \\
    \subfloat[\label{fig:head_c}]{
        \includegraphics[width=0.98\linewidth]{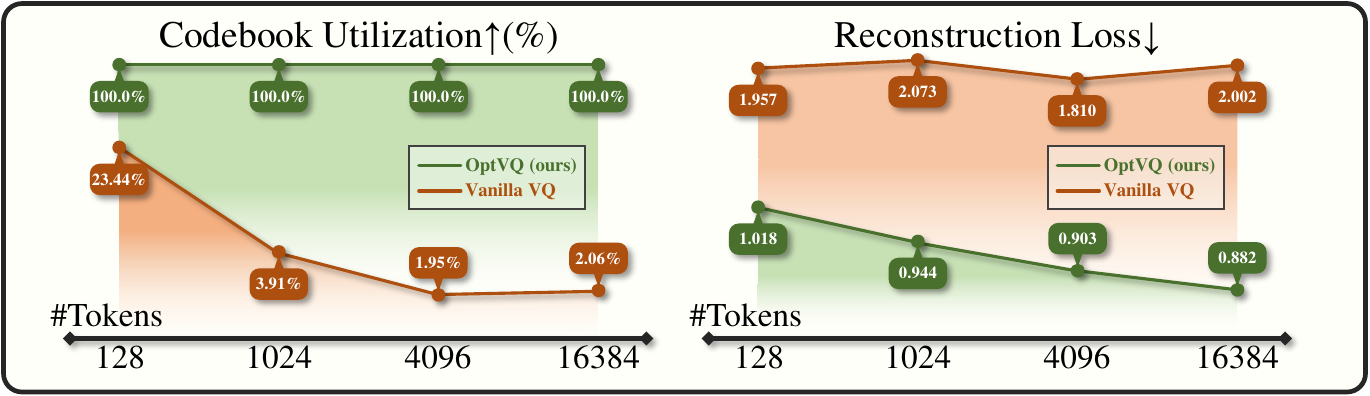}
    }
    \vspace{-2mm}
    \caption{Comparison between different VQ methods. 
    (a) Vanilla VQ employs the nearest neighbor search for quantization, which is a greedy quantization strategy. 
    (b) OptVQ considers vector quantization as an optimal transport problem, which utilizes global information between data for quantization.
    (c) OptVQ achieves 100\% codebook utilization and outperforms other counterparts in image reconstruction tasks.}
    \vspace{-5mm}
\end{figure}

The vector-quantized networks (VQNs), initially proposed for image distribution modeling~\cite{van2017neural}, has since been expanded with enhanced architectures~\cite{esser2021taming,yu2022vector,cao2023efficient} and training strategies~\cite{yu2024image,zhu2024scaling,weber2024maskbit}.
Specifically, the conventional vector quantization calculates the Euclidean distance between data features $\mZ$ and the codebook $\mC$, with each feature selecting the closest codebook token as its discrete representation.
Given the non-backpropagatable nature of the nearest neighbor operation, VQNs approximate the data feature gradient through copying the codebook's gradient during training (i.e.,  straight-through gradient estimation)~\cite{bengio2013estimating}.
Despite their significance, VQNs present training challenges, notably the ``index collapse'' phenomenon.
Conventional VQNs, akin to parameterized online K-Means algorithms~\cite{caron2018deep,huh2023straightening}, are susceptible to local optima.
As depicted in \cref{fig:head_a}, the nearest neighbor operation is a greedy strategy, often leading to the selection of only a few peripheral codebooks, while the majority remain unused. 
To address this, researchers have proposed solutions such as subtle initialization~\cite{huh2023straightening,zhu2024scaling}, distillation strategies~\cite{yu2024image}, and lower-dimensional codebooks~\cite{yu2022vector}.
However, we posit that the training difficulties of VQNs are inherent to the local convergence issues of K-Means algorithms.
A quantization strategy leveraging the global structure of data is essential to evade these local pitfalls.

In this paper, we propose a novel vector quantization strategy, \textbf{OptVQ}, which leverages global structure information between data and codebook for quantization.
Inspired by optimal transport theory~\cite{cuturi2013sinkhorn,asanoself2020,caron2020unsupervised}, we frame vector quantization as an optimal transport problem.
The objective is to learn a mapping from data to the codebook that minimizes the overall transportation cost, as illustrated in \ref{fig:head_b}.
To address this problem with efficiency, we utilize the Sinkhorn-Knopp algorithm~\cite{cuturi2013sinkhorn} to optimize the transport problem.
Theoretical research confirms that the Sinkhorn algorithm can achieve near-optimal assignment results with significant efficiency.
However, it is imperative to recognize that the Sinkhorn algorithm's performance is sensitive to the range of data values.
Our research reveals that a straightforward yet effective normalization technique can mitigate this sensitivity.
Our extensive experiments have demonstrated that OptVQ ensures 100\% codebook utilization and surpasses current state-of-the-art vector quantization methods in image reconstruction tasks. 
OptVQ not only enhances the quality of reconstruction but also achieves training stability without the need for complex training techniques such as subtle initialization or distillation. 
In summary, our contributions are as follows:
\begin{itemize}
    \item \textbf{Optimal transport perspective:} We identify the local pitfalls in VQNs training stability and propose an optimal transport perspective that fully considers the global data structure for quantization.
    \item \textbf{Plug-and-play quantizer OptVQ:} We integrate the Sinkhorn algorithm and develope an efficient quantizer, OptVQ, which achieves 100\% codebook utilization.
    \item \textbf{Techniques to mitigate sensitivity:} By introducing a straightforward normalization technique, we effectively neutralize the impact of varying data ranges.
    \item \textbf{Improved reconstruction ability:} We validated the effectiveness of OptVQ in numerous experiments, proving its superiority in image reconstruction tasks.
\end{itemize}
\vspace{-1mm}
\section{Related Work}
\vspace{-2mm}

The advent of vector quantization~\cite{van2017neural} bridged the gap between continuous and discrete spaces, facilitating the application of VQNs in both image understanding~\cite{baobeit2022,ge2024making,jinunified2024} and generation~\cite{esser2021taming,chang2022maskgit,tian2024visual}.
Despite these applications, VQNs continue to encounter challenges, which we dissect in terms of \textbf{reconstruction performance} and \textbf{training stability}.

\vspace{-3mm}
\paragraph{Reconstruction Performance.}
VQ-VAE~\cite{van2017neural} marked the inception of VQN models, offering a formidable framework for the discretization of continuous data.
However, early VQN models suffered from poor reconstruction quality.
To address this, researchers sought to enhance model capacity through the integration of complex architectures.
VQ-VAE2~\cite{razavi2019generating}, for instance, implemented a multi-scale quantization strategy to retain high-frequency detail.
The leveraging of Vision Transformers~\cite{yu2022vector,yu2024image,cao2023efficient} has also been instrumental in bolstering model capacity.
Concurrently, the efficacy of loss functions was targeted for performance improvements.
VQGAN~\cite{esser2021taming} notably augmented the aesthetic quality of reconstructed images by integrating GANs~\cite{goodfellow2014generative} and perceptual loss~\cite{larsen2016autoencoding,johnson2016perceptual}.
The role of codebooks in reconstruction performance has also been recognized as crucial. 
VQGAN-LC~\cite{zhu2024scaling} demonstrated the benefits of a more extensive codebook in bolstering reconstruction capabilities.
MoVQ~\cite{zheng2022movq} and RQ-VAE~\cite{lee2022autoregressive,tian2024visual} have respectively introduced multi-head and residual mechanisms to increase the equivalent codebook size without a increase in actual size.
Meanwhile, MAGVIT-v2~\cite{yulanguage} and MaskBit~\cite{weber2024maskbit} have proposed lookup-free and embedding-free methodologies to enhance reconstruction efficacy.

\vspace{-3mm}
\paragraph{Training Stability.}
The stability of VQN training is a subject of ongoing debate, with the phenomenon of ``index collapse'' being a prevalent challenge.
To address this, various strategies have been proposed to enhance training robustness, 
including low-dimensional codebooks~\cite{yu2022vector}, shared affine transformations~\cite{huh2023straightening,zhu2024scaling}, specialized initializations~\cite{huh2023straightening,zhu2024scaling}, and model distillation~\cite{yu2024image}. 
ViT-VQGAN~\cite{yu2022vector} observed that in high-dimensional spaces, the feature space is notably sparse, and reducing the dimensionality of the codebook can increase the proximity of the codebook to features, thereby improving utilization.
Shared affine transformations~\cite{huh2023straightening,zhu2024scaling} have demonstrated that the conventional codebook update method is sparse and slow, inadequate for keeping pace with feature evolution.
An affine transformation layer after the codebook allows for the conversion of sparse updates into dense ones, thus enhancing the update process.
To prevent arbitrary initializations from leading to premature convergence, K-Means initialization~\cite{huh2023straightening,zhu2024scaling} has been identified as a reliable approach.
Additionally, distillation from a well-trained VQN, such as MaskGiT~\cite{chang2022maskgit}, has been shown to improve performance~\cite{yu2024image}.
These cumulative insights into stability prompt the consideration of a more fundamental and straightforward approach to VQN training stability.
The reliance on nearest neighbor search in conventional vector quantization, akin to the properties of online K-Means~\cite{caron2018deep,huh2023straightening}, is identified as the primary cause of susceptibility to local optima or index collapse.
A comprehensive solution to this challenge may lie in the adoption of a search methodology that incorporates global structural awareness, leading us to explore optimal transport as a potential solution.
\section{Method}

In this section, we initially outline the foundational concepts of vector-quantized networks and the training procedures in \cref{sec:preliminary}.
Subsequently, we propose \textbf{OptVQ}, which leverages optimal transport to replace the nearest neighbor search, thereby achieving a global-aware vector quantization in \cref{sec:optvq}.
Finally, we explore techniques to mitigate the sensitivity of the optimal transport algorithm and present the refined OptVQ algorithm in \cref{sec:technique}.

\subsection{Preliminaries: Vector-Quantized Networks} \label{sec:preliminary}

The pursuit of compressed representations is important in various applications.
Pioneering work was conducted by Autoencoders~\cite{hinton2006reducing}, which were the first to encode images into a lower-dimensional space.
The VAE~\cite{kingma2013auto} further advanced this concept by introducing a probabilistic framework, 
constraining the encoded representations to follow a specific distribution (e.g., Gaussian), thereby enabling data generation through sampling. 
These initial methods predominantly focused on obtaining continuous compressed representations.
However, with the rise of large-scale sequence prediction models~\cite{brown2020language,achiam2023gpt,touvron2023llama}, the interest in discrete data representations has surged.
The Vector-Quantized VAE (VQ-VAE)~\cite{van2017neural} addressed this by replacing the continuous prior distribution with a discrete one. 
This paradigm has since gained widespread adoption in image generation~\cite{esser2021taming,chang2022maskgit,rombach2022high} and large-scale pre-training~\cite{baobeit2022,bai2024sequential}.

\paragraph{Mathematical Notation.}
Let the dataset be denoted as $\mathcal{X} = \{\vx_i\}_{i=1}^N$.
A typical vector-quantized network (VQN) comprises three main components: an encoder $f(\cdot)$, a decoder $g(\cdot)$, and a codebook $h(\cdot)$.
The encoder $f$ maps the input data $\vx$ to a feature tensor $\mZ_e = f(\vx) \in \mathbb{R}^{m \times d}$,
and we denote the vector in $\mZ_e$ as $\vz_e \in \mathbb{R}^d$.
The quantizer $h$ then selects the closest code vector from the codebook $\mathcal{C} = \{c_1, c_2, \ldots, c_n\}$ to obtain the quantized vectors $\vz_q = h(\vz_e, \mathcal{C})$,
which constitute the quantized tensor $\mZ_q$.
For simplicity, we will omit the notation $\mathcal{C}$ in the quantizer $h$ hereafter.
Each $c_i$ is referred to as a code vector, with the index $i$ being the code or token.
The decoder $g$ subsequently reconstructs the original data from the quantized tensor $\mZ_q$, yielding $\hat{\vx} = g(\mZ_q)$.

\paragraph{Quantization Operation.}
The quantizer $h$ is commonly implemented via a nearest neighbor search~\cite{van2017neural,razavi2019generating,esser2021taming,chang2022maskgit,yu2022vector}, which is formalized as follows:
\begin{align} \label{equ:nearest}
    \vz_q = h(\vz_e) = \vc_k, ~ \text{where} ~ k = \argmin_{j} \Vert \vz_e - \vc_j \Vert,
\end{align}
with $\Vert \cdot \Vert$ representing any distance metric, such as the Euclidean distance.
For image data $\vx \in \mathbb{R}^{h_I \times w_I \times 3}$, the encoder $f$ typically extracts a feature tensor $\mZ_e \in \mathbb{R}^{h \times w \times d}$, where the downsampling ratio is $r = h_I / h = w_I / w$.
The quantizer $h$ is applied to each spatial location of the tensor, such that $\mZ_q[i,j] = h(\mZ_e[i,j]) \in \mathbb{R}^d$.

\paragraph{Training Objective.}
The training of VQNs is conventionally performed by minimizing the reconstruction loss~\cite{van2017neural}, denoted as $\mathcal{L}_{\text{rec}} = \Vert \vx - \hat{\vx} \Vert$.
While different norms are theoretically equivalent in finite-dimensional spaces (e.g., $l_2 \leq l_1 \leq \sqrt{d} ~ l_2$), 
the $l_1$-based reconstruction loss often imposes stricter constraints than the $l_2$-based one from an optimization perspective, as indicated by studies on exact penalty function methods~\cite{han1979exact,di1994exact}.
Consequently, subsequent research advocates the $l_1$ norm for reconstruction constraints.
Moreover, since the reconstruction function may overemphasize low-level details and not align perfectly with human perception~\cite{larsen2016autoencoding,johnson2016perceptual,dosovitskiy2016generating}, 
some studies enhance the aesthetic appeal of image reconstruction by incorporating perceptual and adversarial losses~\cite{esser2021taming,chang2022maskgit,yu2022vector,cao2023efficient}.
Thus, a composite loss function is often employed in VQN training.
Empirically, adversarial loss has the most significant impact on aesthetics, followed by perceptual loss, $l_1$-based reconstruction loss, and $l_2$-based one.

\paragraph{Straight-Through Estimator.}
Calculating gradients for VQNs directly using the chain rule~\cite{rumelhart1986learning} is challenging due to the non-differentiability of the nearest neighbor search in \cref{equ:nearest}.
This non-differentiability precludes the computation of gradients for the encoder parameters $\theta_f$.
To address this, the straight-through estimator~\cite{bengio2013estimating,huh2023straightening} is utilized, which approximates the operation as:
\begin{align} \label{equ:ste}
    \vz_q = \vz_e + \text{sg}(\vz_q - \vz_e),
\end{align}
where $\text{sg}(\cdot)$ denotes the stop-gradient operation.
This estimator approximates the derivative $\frac{\partial \vz_q}{\partial \vz_e}$ to $1$.
Consequently, the gradient of the loss $\mathcal{L}$ with respect to the encoder parameters $\theta_f$ can be approximated by:
\begin{align} \label{equ:chain_rule}
    \nabla_{\theta_f} \mathcal{L} = 
    \frac{\partial \mathcal{L}}{\partial \hat{\vx}}
    \frac{\partial \hat{\vx}}{\partial \vz_q}
    \cancel{\frac{\partial \vz_q}{\partial \vz_e}}
    \frac{\partial \vz_e}{\partial \theta_f}
    \approx \frac{\partial \mathcal{L}}{\partial \hat{\vx}}
    \frac{\partial \hat{\vx}}{\partial \vz_q}
    \frac{\partial \vz_e}{\partial \theta_f} 
    = \hat{\nabla}_{\theta_f} \mathcal{L}.
\end{align}
To reduce the approximation error between $\nabla{\theta_f} \mathcal{L}$ and $\hat{\nabla}{\theta_f} \mathcal{L}$, a commitment loss~\cite{van2017neural} is introduced to align the features $\vz_e$ with their quantized counterparts $\vz_q$:
\begin{align}
    \mathcal{L}_{\text{cmt}} = \Vert sg(\vz_e) - \vz_q \Vert + \beta \Vert \vz_e - sg(\vz_q)\Vert,
\end{align}
where $\beta$ is a hyperparameter balancing the two terms. 
The overall training objective for VQNs is thus:
\begin{align} \label{equ:vqn_loss}
    \mathcal{L} = 
    \underbrace{\Vert \vx - \hat{\vx} \Vert}_{\text{Recon. Loss}} +
    \underbrace{\Vert sg(\vz_e) - \vz_q \Vert + \beta \Vert \vz_e - sg(\vz_q)\Vert}_{\text{Commit. Loss}}.
\end{align}

\subsection{OptVQ: Global-Aware Tokenization} \label{sec:optvq}

\begin{figure}[tbp]
    \centering
    \subfloat[Vanilla VQ\label{fig:process_a}]{
        \includegraphics[width=0.98\linewidth]{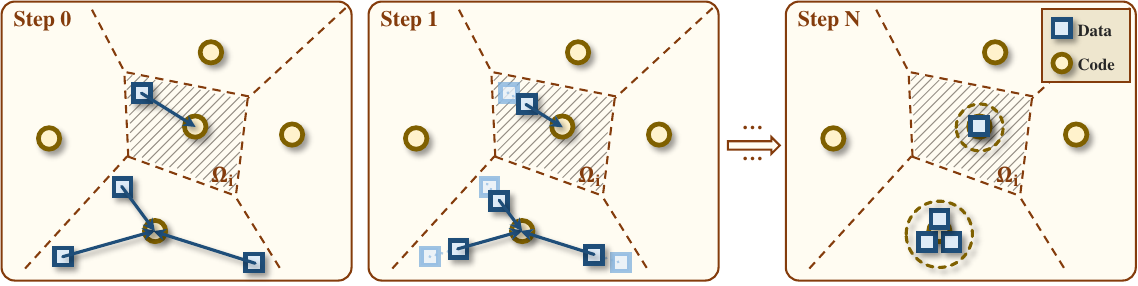}
    }\\
    \subfloat[OptVQ \small{(ours)}\label{fig:process_b}]{
        \includegraphics[width=0.98\linewidth]{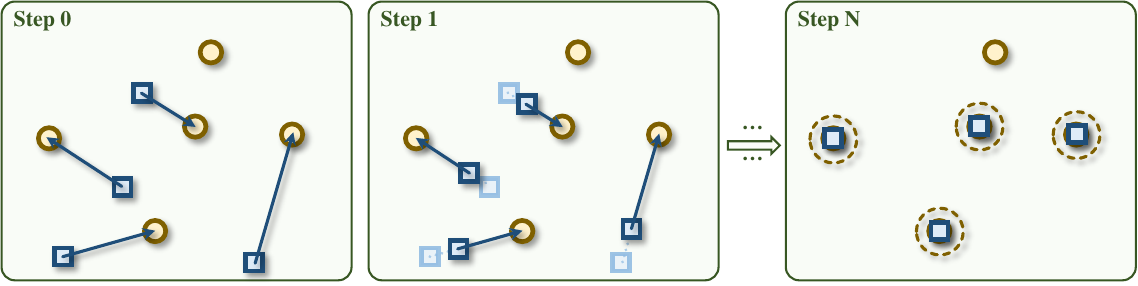}
    }
    \vspace{-2mm}
    \caption{Optimization process of different quantization methods. (a) Vanilla VQ are significantly impacted by initialization, and features are trapped in the Voronoi cell $\Omega_i$. (b) The proposed OptVQ can escape local dilemmas and achieve global-aware indexing.}
    \vspace{-4mm}
\end{figure}

\paragraph{Local Pitfalls of VQ.}
While VQ-VAEs have achieved substantial success across various domains, their training instability has become increasingly evident, particularly with the occurrence of ``index collapse'' issues~\cite{yu2022vector,huh2023straightening,zhu2024scaling,yu2024image}.
This phenomenon is attributed to the local nature of the nearest neighbor search, inherent to the K-Means algorithm~\cite{caron2018deep, huh2023straightening}.
The initialization of the codebook $\mathcal{C}$ is pivotal as it directly influences the assignment of codes to features by nearest neighbor search, which is challenging to rectify in subsequent iterations.
Specifically, if the codebook distribution is misaligned with the features during  initialization, it can lead to a scenario where only a minority of codebooks are utilized throughout the training process.
As illustrated in \cref{fig:process_a}, a feature vector $\vz_k \in \mathbb{R}^d$ is assigned to the code $\vc_i$ in the codebook $\mathcal{C}$ if the distance $\Vert \vz_k - \vc_i \Vert$ is the smallest among all code vectors:
\begin{align} \label{equ:nearest_cond}
    \Vert \vz_k - \vc_i \Vert < \Vert \vz_k - \vc_j \Vert, ~ \forall j \neq i.
\end{align}
By simplifying this inequality, we can deduce that $\vz_k$ is enclosed within the Voronoi cell $\Omega_i$ of the code $\vc_i$:
\begin{align} \label{equ:voronoi}
    (\vc_j - \vc_i)^T \vz_k < \Vert \vc_j \Vert^2_2 - \Vert \vc_i \Vert^2_2, \quad \forall j \neq i.
\end{align}
Given that Voronoi cells are intersections of multiple half-planes, they are convex sets.
To simplify our analysis, we assume that the codebook evolves slowly and focus primarily on the impact of the commitment loss $\mathcal{L}_{\text{cmt}}$ on the distribution.
We demonstrate that if $\vz_k \in \Omega_i$ at time $t$, then post-update at time $t+1$, $\vz_k$ remains within $\Omega_i$. 
Firstly, the updating direction for $\vz_k$ is given by $\Delta \vz_k = \gamma (\vc_i - \vz_k)$, where $\gamma$ represents the learning rate, typically a small value.
After updating, we obtain $\vz_k' = \vz_k + \gamma (\vc_i - \vz_k)$.
Clearly, when $\gamma$ is small, $\vz_k'$ lies on the line segment between $\vz_k$ and $\vc_i$.
By the properties of convex sets, it follows that $\vz_k'$ is still trapped in $\Omega_i$.
Hence, the initialization of the codebook is of paramount importance.
Some studies propose the use of specialized initialization methods to position the codebook closer to the data distribution, thereby mitigating the occurrence of index collapse~\cite{huh2023straightening, zhu2024scaling}.
However, as the training progresses, the codebook may diverge from the data distribution, indicating that this problem cannot be solely resolved through initialization.

\vspace{-3mm}
\paragraph{Global Assignment via Optimal Transport.}
We propose that a fundamental solution to the aforementioned problem lies in employing a global search method instead of the nearest neighbor search, as depicted in \cref{fig:process_b}.
Optimal transport theory~\cite{cuturi2013sinkhorn, asanoself2020, caron2020unsupervised} is dedicated to solving the optimal mapping problem between two distributions. 
The core concept is that by minimizing a certain cost, we can map one distribution onto another, effectively addressing our local pitfall issue in VQ.
Specifically, for the codebook $\mC = [\vc_1, \ldots, \vc_n] \in \mathbb{R}^{d \times n}$ and the features to be assigned $\mZ = [\vz_1, \ldots, \vz_l] \in \mathbb{R}^{d \times l}$,
we seek an assignment matrix $\mA \in \mathbb{R}_+^{l \times n}$ that not only respects the order of distances between codes and features, but also ensures maximal participation of each code and feature. 
This can be formulated as the following optimization problem:
\begin{align} \label{equ:optimal_transport}
    &\min_{\mA} \text{Tr}(\mA^T \mD) - \frac{1}{\epsilon} H(\mA) \\
    &s.t., ~ \mA \bm{1}_r = \bm{1}_r, \mA^T \bm{1}_c = \bm{1}_c, \mA \in \mathbb{R}_+^{l \times n}, \notag
\end{align}
where $\mD \in \mathbb{R}_+^{l \times n} (\mD_{ij} = \Vert \vz_i - \vc_j \Vert)$ is the distance matrix between codes and features, $H(\mA) = - \sum_{ij} \mA_{ij} \log \mA_{ij}$ is the entropy function, $\epsilon$ is a hyperparameter, and $\bm{1}_r$ and $\bm{1}_c$ are vectors of ones for the rows and columns, respectively.
The objective function $\text{Tr}(\mA^T \mD)$ ensures that the assignment matrix $\mA$ adheres to the order of distances between codes and features, while the entropy term $H(\mA)$ and the constraints ensure that all codes and features are considered in the assignment process.
By selecting an appropriate value for $\epsilon$, a good balance can be achieved between maintaining the order of distances and ensuring a globally informed assignment.
The larger the value of $\mA_{ij}$, the greater the tendency for feature $\vz_i$ to be assigned to code $\vc_j$. 
Thus, the OptVQ algorithm we propose is as follows:
\begin{align} \label{equ:optvq}
    \vz_q = h(\vz_i) = \vc_k, \quad \text{where} \quad k = \argmax_{j} \mA_{ij}.
\end{align}
When the data distribution is significantly inconsistent with the codebook distribution, the entropy function and constraint term in OptVQ play a crucial role in ensuring that each code and feature participate fully in the assignment, thereby avoiding the local pitfall issue.
As training progresses and the distributions of data and codebook gradually overlap (i.e., there exists a code $\vc_j$ such that the distance to feature $\vz_i$ is close to $0$), the solution to \cref{equ:optimal_transport} will tend towards a one-hot format, where $\mA_{ij} > 0$ and $\mA_{ik} = 0$ for all $k\neq j$, in which OptVQ aligns with the conventional VQ method.
Subsequent experiments verify that our proposed OptVQ effectively solves the local pitfalls of conventional VQ and basically achieves 100\% codebook utilization.

\subsection{Robust Training of OptVQ} \label{sec:technique}

This section elucidates the integration of the Sinkhorn-Knopp algorithm to efficiently address the optimization problem presented in \cref{equ:optimal_transport}, with low computational demands. 
We also introduce a straightforward yet robust normalization technique to attenuate the sensitivity of the Sinkhorn algorithm to the range of input data. 

\begin{figure}[tbp]
    \centering
    \subfloat[Sinkhorn algorithm\label{fig:tech_a}]{
        \includegraphics[width=0.48\linewidth]{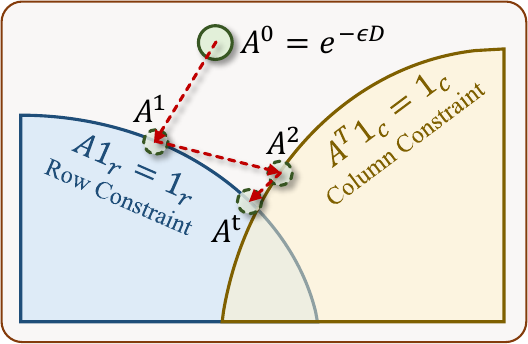}
    }
    \subfloat[Numerical sensitivity\label{fig:tech_b}]{
        \includegraphics[width=0.48\linewidth]{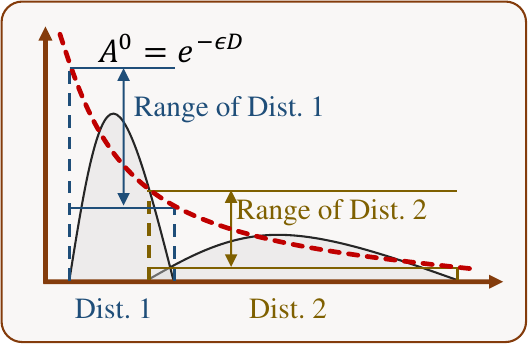}
    }\\
    \subfloat[Multi-head quantizer\label{fig:tech_c}]{
        \includegraphics[width=0.98\linewidth]{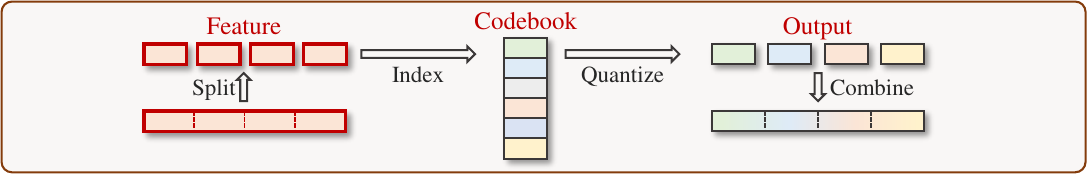}
    }
    \vspace{-2mm}
    \caption{Training details of OptVQ. (a) The iterative Sinkhorn algorithm efficiently resolves the optimal transport problem. (b) The original Sinkhorn method exhibits sensitivity to input values, necessitating normalization. (c) The multi-head mechanism significantly amplifies the effective number of codebooks.}
    \vspace{-3mm}
\end{figure}

\vspace{-3mm}
\paragraph{Sinkhorn-Knopp Algorithm.}
While \cref{equ:optimal_transport} is a convex problem, solving it directly using conventional algorithms such as gradient descent requires significant computational effort. 
The Sinkhorn-Knopp algorithm, as proposed in optimal transport theory~\cite{cuturi2013sinkhorn}, offers an efficient alternative, yielding approximate solutions through a limited number of iterations, as illustrated in \cref{fig:tech_a}. 
This algorithm has been widely adopted in feature clustering~\cite{asanoself2020} and self-supervised learning~\cite{caron2020unsupervised}.
The Sinkhorn algorithm initiates with:
\begin{align} \label{equ:sinkhorn}
    \mA^0 = e^{- \epsilon \mD}.
\end{align}
Subsequently, it alternates between normalizing the rows and columns of $\mA^t$ until convergence as follows:
\begin{align}
    \mA^{t+1}_{ij} &= \mA^t_{ij} / \sum_k \mA^t_{ik}, \tag{Row Norm.} \\
    \mA^{t+2}_{ij} &= \mA^{t+1}_{ij} / \sum_k \mA^{t+1}_{kj}. \tag{Column Norm.}
\end{align}
Our experiments indicate that no more than $5$ iterations are enough to achieve a solution proximate to the optimal.

\vspace{-3mm}
\paragraph{Normalization Technique.}
The Sinkhorn algorithm, as delineated in \cref{equ:sinkhorn}, employs an exponential form for initialization.
A large value in the input variable $\mD$ can lead to elements of $\mA$ approaching zero, potentially inducing numerical instability due to computational precision limits, as depicted in \cref{fig:tech_b}. 
Drawing upon normalization techniques to manage distribution variations (e.g., Batch-Normalization~\cite{ioffe2015batch}), we normalized the input $\mD$ as follows:
\begin{align} 
    \label{equ:normalization_1}
    \mD'_{ij} &= \frac{\mD_{ij} - \bar{\mD}_{ij}}{\text{std}(\mD_{ij})} \\
    \label{equ:normalization_2}
    \mD''_{ij} &= \mD'_{ij} - \min(\mD'_{ij}).
\end{align}
Here, the decentration operation in \cref{equ:normalization_1} standardizes the value range, 
and the non-negativization operation in \cref{equ:normalization_2} constrains the lower bound of $\mD_{ij}$, preventing excessively small values.
Such small values could cause $\mA_{ij} = e^{-\epsilon \mD_{ij}}$ to approach infinity, leading to numerical issues.
These normalization steps effectively decouple the selection of $\epsilon$ from the range of $\mD$ values, simplifying the determination of $\epsilon$ and enhancing the algorithm's robustness against variations in input data scales.

\vspace{-3mm}
\paragraph{Multi-head Quantizer.}
The size of codebooks is a pivotal factor in the reconstruction performance of VQNs. 
Some studies have attempted to enhance performance by simply increasing the number of codebooks~\cite{zhu2024scaling}. 
However, as per rate-distortion theory~\cite{shannon1959coding}, improving reconstruction performance may necessitate an exponential increase in codebooks, which poses a substantial computational challenge.
We implement the multi-head quantization proposed in MoVQ~\cite{zheng2022movq}, as shown in \cref{fig:tech_c}. 
Assuming $B$ heads are utilized, the code dimension in the codebook is adjusted to $d/B$. 
Initially, $\vz_e$ is divided into $B$ segments, 
each quantized independently through the OptVQ algorithm, 
and the quantized segments are concatenated to form the final quantized vector, 
implying that each feature vector is represented by $B$ tokens. 
The advantage of this scheme lies in the cross-combination of $n$ codebooks through the Cartesian product, effectively amplifying the equivalent codebook size to $n^B$.
\begin{table*}[tbp]
    \centering
    \caption{Quantitative comparison with state-of-the-art methods on ImageNet.}
    \vspace{-2mm}
    \begin{tabular}{|c|c|c|c|cccc|}
    \hline
    \textbf{Model} & \textbf{Latent Size} 
    & \textbf{From Scratch} & \textbf{\#Tokens} 
    & \textbf{SSIM↑} & \textbf{PSNR↑} & \textbf{LPIPS↓} & \textbf{rFID↓} \bigstrut\\
    \hline
    taming-VQGAN~\cite{esser2021taming} & $16\times 16$ & \CheckmarkBold 
        & 1,024  & 0.521  & 23.30  & 0.195  & 6.25  \bigstrut[t]\\
    MaskGIT-VQGAN~\cite{chang2022maskgit} & $16\times 16$ & \CheckmarkBold 
        & 1,024  & -     & -     & -     & 2.28  \\
    Mo-VQGAN~\cite{zheng2022movq} & $16\times 16\times 4$ & \CheckmarkBold 
        & 1,024  & 0.673  & 22.42  & 0.113  & 1.12  \\
    TiTok-S-128~\cite{yu2024image} & $128$ & \XSolidBrush 
        & 4,096  & -     & -     & -     & 1.71  \\
    ViT-VQGAN~\cite{yu2022vector} & $32\times 32$ & \CheckmarkBold 
        & 8,192  & -     & -     & -     & 1.28  \\
    taming-VQGAN~\cite{esser2021taming} & $16\times 16$ & \CheckmarkBold 
        & 16,384 & 0.542  & 19.93  & 0.177  & 3.64  \\
    RQ-VAE~\cite{lee2022autoregressive} & $8\times 8\times 16$ & \CheckmarkBold 
        & 16,384 & -     & -     & -     & 1.83  \\
    VQGAN-LC~\cite{zhu2024scaling} & $16\times 16$ & \XSolidBrush 
        & 100,000 & 0.589  & 23.80  & 0.120  & 2.62  \bigstrut[b]\\
    \hline
    \multirow{2}[2]{*}{\textbf{OptVQ (ours)}} & \textbf{$\bm{16\times 16\times 4}$} & \CheckmarkBold 
        & \textbf{16,384} & \textbf{0.717}  & \textbf{26.59}  & \textbf{0.076}  & \textbf{1.00}  \bigstrut[t]\\
        & \trc{$\bm{16\times 16\times 8}$} & \trc{\CheckmarkBold} 
        & \trc{16,384} & \trc{0.729}  & \trc{27.57}  & \trc{0.066}  & \trc{0.91}  \bigstrut[b]\\
    \hline
    \end{tabular}
    \label{tab:exp_recon_imagenet}
\end{table*}

\begin{figure*}[tbp]
    \centering
    \includegraphics[width=0.95\linewidth]{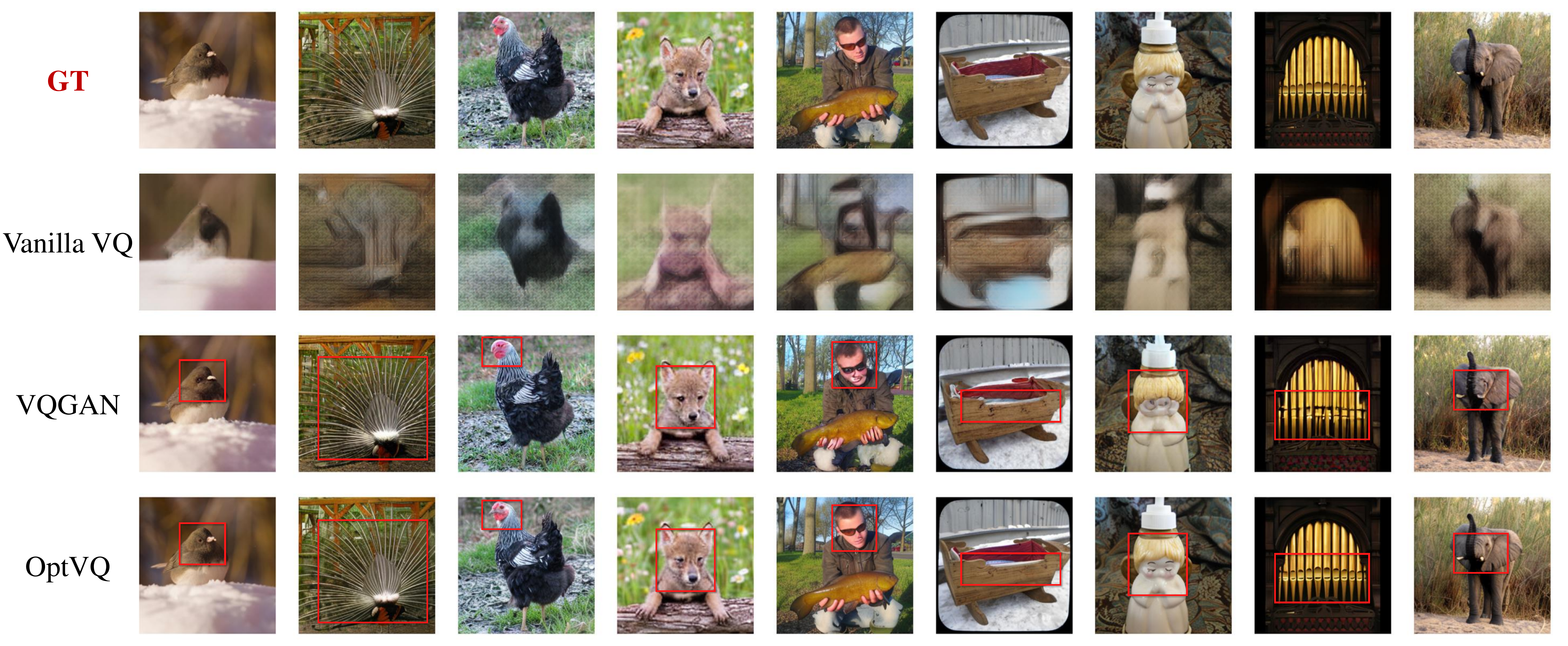}
    \vspace{-2mm}
    \caption{Visualizations of reconstruction results on ImageNet validation set (detailed comparison marked in \trc{red} boxes).}
    \vspace{-3mm}
    \label{fig:exp_recon_imagenet}
\end{figure*}

\begin{table}[tbp]
    \centering
    \caption{Quantitative comparison on MNIST and CIFAR-10.}
    \begin{tabular}{|c|c|cc|}
    \hline
    \textbf{Dataset} & \textbf{Model} & \textbf{PSNR↑} & \textbf{Code Usage↑} \bigstrut\\
    \hline
    \multirow{2}[2]{*}{\textbf{MNIST}} 
    & VQ-VAE & 9.02  & 0.20\% \bigstrut[t]\\
    & \trc{OptVQ (ours)} & \trc{34.28}  & \trc{100.00\%} \bigstrut[b]\\
    \hline
    \multirow{2}[2]{*}{\textbf{CIFAR-10}} 
    & VQ-VAE & 14.57  & 0.10\% \bigstrut[t]\\
    & \trc{OptVQ (ours)} & \trc{28.78}  & \trc{100.00\%} \bigstrut[b]\\
    \hline
    \end{tabular}
    \vspace{-2mm}
    \label{tab:exp_recon_small}
\end{table}

\vspace{-3mm}
\section{Experiments}
\vspace{-2mm}

In this section, we present a comprehensive experiments to substantiate the effectiveness of OptVQ.
We begin by comparing the reconstruction performance across three prominent datasets in \cref{sec:exp_rec}.
This is followed by a series of meticulously crafted experiments to validate the enhancements that OptVQ offers to overcome local optima in \cref{sec:exp_veri}.
Finally, we engage in ablation studies to evaluate the influence of optimal transport operations, normalization techniques, and parameter selection on the overall performance in \cref{sec:exp_abl}.

\subsection{Experiment Details}

\paragraph{Model Settings.}
The models and loss functions are based on the VQGAN~\cite{esser2021taming}.
In terms of model architecture, the autoencoder in our experiments accepts input images with a resolution of $256 \times 256$ pixels and employs a downsampling factor of $f=16$.
For the loss function, we adopted a hybrid approach for image reconstruction tasks, consistent with VQGAN, which encompasses perceptual loss, $l_1$-based loss, and $l_2$-based loss. 
For vector quantization tasks, we incorporated commitment loss. 
\begin{figure}[tbp]
    \centering
    \includegraphics[width=\linewidth]{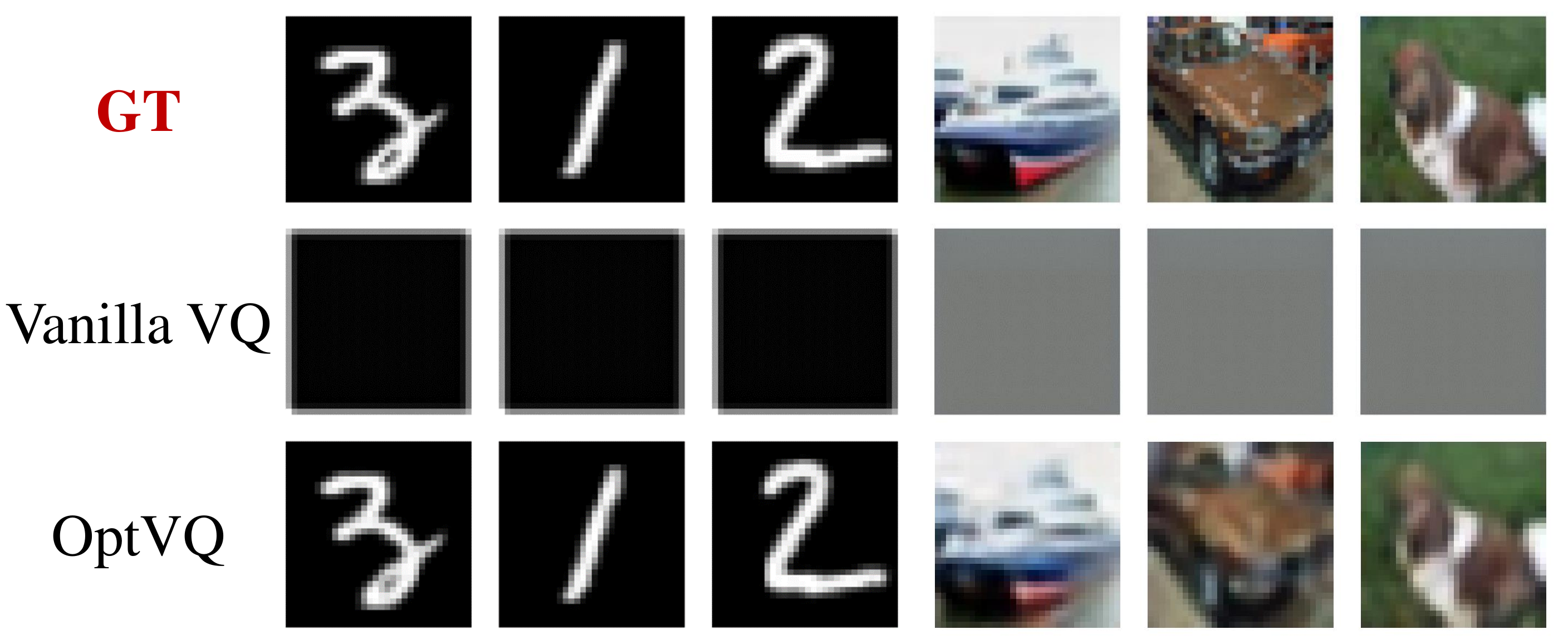}
    \vspace{-2mm}
    \caption{Visualizations on MNIST and CIFAR-10.}
    \vspace{-2mm}
    \label{fig:exp_recon_small}
\end{figure}
For adversarial training, we leveraged the loss function associated with patch-based discriminators~\cite{isola2017image}.
These loss functions are amalgamated through a weighted summation, where the perceptual loss is assigned a weight of $3$, and all other losses are weighted equally at $1$.
Unless specified otherwise, the quantizer parameters are configured as follows:
the codebook size is set to $n=16384$,
the feature dimension is $d=64$,
a single fully connected layer serves as the shared affine transformation for the codebook~\cite{huh2023straightening},
the multi-head quantization mechanism parameter is $B=4$,
the commitment loss weight is $\beta=0.25$,
and the Sinkhorn algorithm is parameterized with a balance factor $\epsilon=10$ and iterated $5$ times.

\begin{figure*}[tbp]
    \centering
    \includegraphics[width=\linewidth]{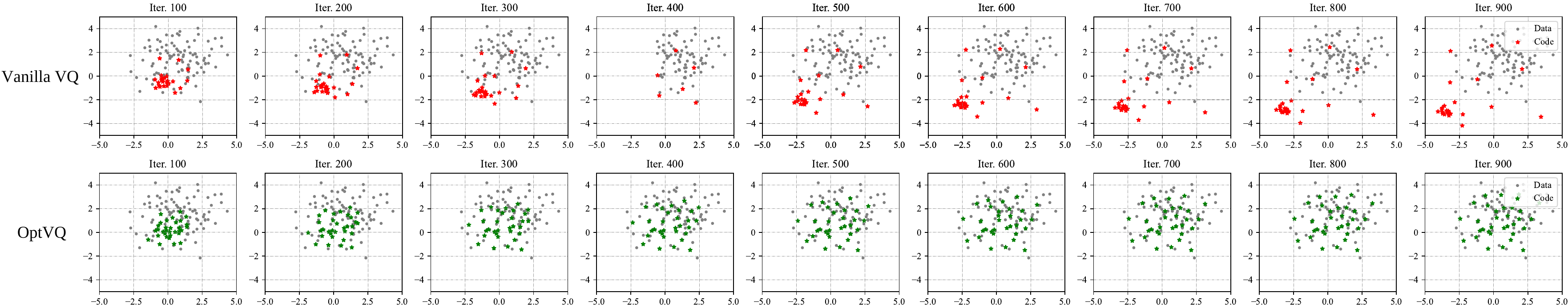}
    \caption{The dynamics of the quantization operation in the two-dimensional case.
    Data points are represented in gray, the codebooks of the nearest neighbor method are indicated in \trc{red}, and the codebooks of our OptVQ method are depicted in \tgc{green}.}
    \vspace{-4mm}
    \label{fig:exp_process}
\end{figure*}

\paragraph{Optimizer Settings.}
The implementation of all models was facilitated by the PyTorch framework~\cite{paszke2019pytorch}, 
and training was conducted on a single machine equipped with eight NVIDIA 4090 GPUs.
We selected AdamW~\cite{loshchilov2019decoupled} as the optimizer. 
Constrained by GPU memory limitations, we configured the batch size per GPU to $8$, culminating in a total batch size of $64$. 
The learning rate was calibrated to $64 \times (2 \times 10^{-6}) = 1.28 \times 10^{-4}$, 
employing the OneCycle learning rate scheduling policy~\cite{smith2019super}. 
The models underwent training for $50$ epochs, with the entire training regimen spanning approximately 8 days.

\subsection{Reconstruction Performance} \label{sec:exp_rec}

To substantiate the effectiveness of OptVQ, an initial evaluation is conducted on MNIST~\cite{lecun1998gradient} and CIFAR-10~\cite{krizhevsky2009learning}.
All images are resized to $32 \times 32$ pixels.
The encoder and decoder are simple 5-layer CNNs, with a downsampling factor of $f=4$.
To maintain comparability, the codebook size is uniformly set to $1024$, with a feature dimension of $8$.
No specialized techniques are employed, such as the subtle initialization~\cite{huh2023straightening,zhu2024scaling}, model distillation~\cite{yu2024image}, or shared affine transformation~\cite{huh2023straightening}.
\cref{tab:exp_recon_small} illustrates that OptVQ can attain 100\% codebook utilization.
This outcome underscores the significant mitigation of the prevalent ``index collapse'' issue in VQNs,
leading to a marked enhancement in reconstruction performance.
The qualitative results in \cref{fig:exp_recon_small} further highlights that OptVQ avoids the collapse phenomenon and produces high-quality reconstruction.

We conduct a comparative analysis of our OptVQ method against leading VQNs on ImageNet~\cite{deng2009imagenet}.
We evaluate the preformance using the PSNR, SSIM~\cite{wang2004image}, LPIPS~\cite{zhang2018unreasonable}, and rFID~\cite{heusel2017gans} metrics on the ImageNet validation set, as presented in \cref{tab:exp_recon_imagenet}. 
Our implementation employs the widely adopted VQGAN architecture~\cite{esser2021taming}, with the codebook size set to $16384$ and the feature dimension to $64$.
The quantitative  results demonstrate that our OptVQ ourperforms its counterparts across all metrics.
Further, \cref{fig:exp_recon_imagenet} elucidates the comparative visual fidelity, with the red-boxed regions underscoring the detail preservation of our method, particularly in facial features and texture information, when compared with other state-of-the-art methods. 

\begin{figure}[tbp]
    \centering
    \includegraphics[width=0.9\linewidth]{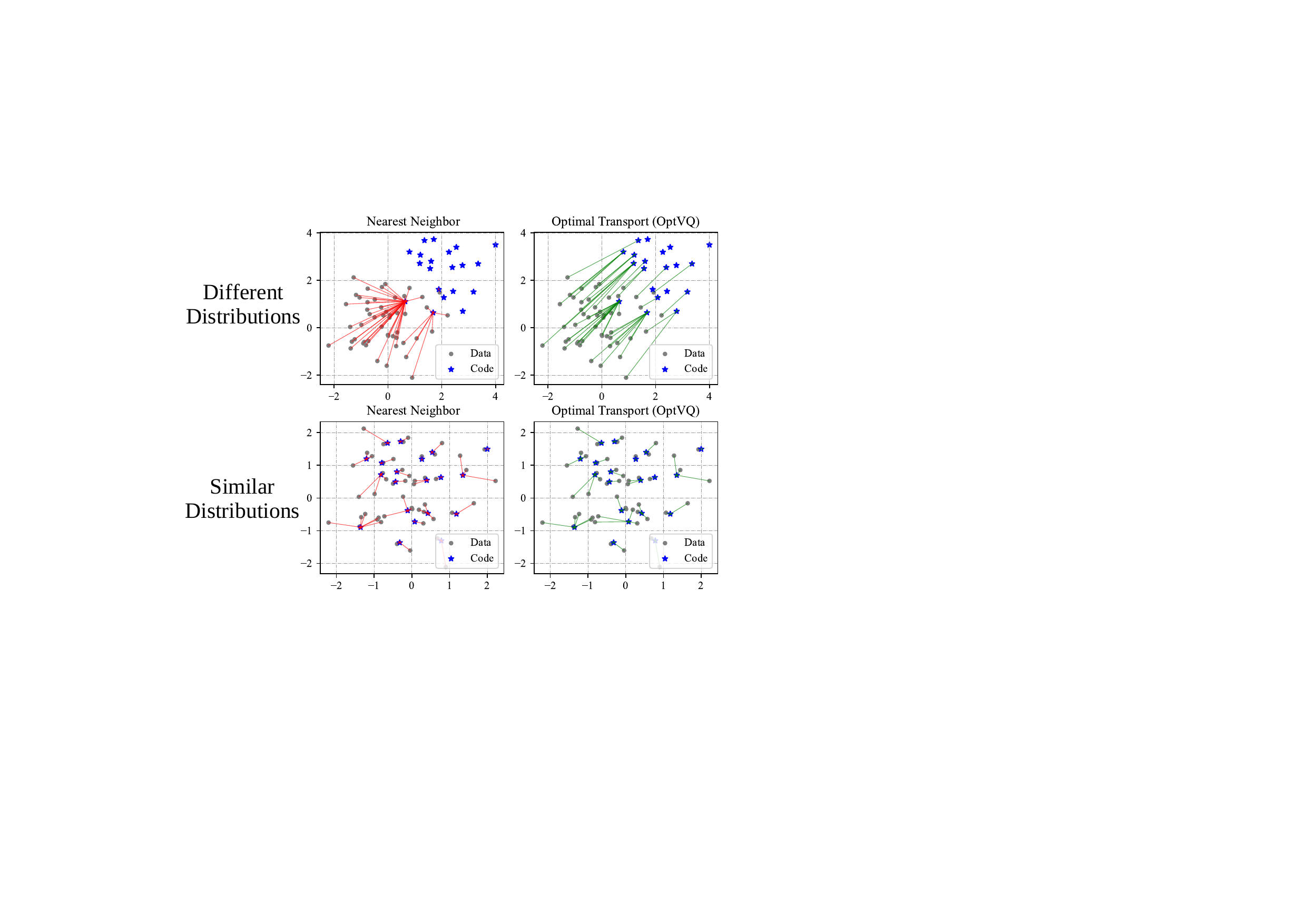}
    \caption{Consistency check between OptVQ and vanilla VQ. \trc{Red} and \tgc{green} arrows are employed to denote the mapping between data points and codes for vanilla VQ and OptVQ, respectively.}
    \vspace{-3mm}
    \label{fig:exp_consistency}
\end{figure}

\begin{table*}[tbp]
    \centering
    \caption{The impact on codebook utilization and reconstruction performance under various conditions.}
    \begin{tabular}{|c|c|cccc|cccc|}
    \hline
    \multirow{2}[4]{*}{\textbf{Metric}} & \multirow{2}[4]{*}{\textbf{\#Tokens}} 
    & \multicolumn{4}{c|}{\textbf{Latent Dim. = 64}} 
    & \multicolumn{4}{c|}{\textbf{Latent Dim. = 8}} \bigstrut\\
    \cline{3-10} & & \textbf{128} & \textbf{1024} & \textbf{4096} & \textbf{16384} & \textbf{128} & \textbf{1024} & \textbf{4096} & \textbf{16384} \bigstrut\\
    \hline
    \multirow{2}[2]{*}{\boldmath{}\textbf{$L_{\text{rec}}$↓}\unboldmath{}} 
    & VQ-VAE & 1.957 & 2.073  & 1.810  & 2.002  & 2.254  & 1.916  & 2.009  & 2.099 \bigstrut[t]\\
    & OptVQ & 1.018  & 0.944  & 0.903  & 0.882  & 0.971  & 0.848  & 0.808  & 0.805 \bigstrut[b]\\
    \hline
    \multirow{2}[2]{*}{\textbf{Code Usage↑}} 
    & VQ-VAE & 23.44\% & 3.91\% & 1.95\% & 2.06\% & 18.75\% & 2.93\% & 1.71\% & 0.20\% \bigstrut[t]\\
    & OptVQ & 100\% & 100\% & 100\% & 100\% & 100\% & 100\% & 100\% & 100\% \bigstrut[b]\\
    \hline
    \end{tabular}
    \vspace{-3mm}
    \label{tab:exp_abl_sinkhorn}
\end{table*}

\subsection{Property Verification} \label{sec:exp_veri}

In this section, we delve into the global-awareness property of OptVQ, which is pivotal in overcoming local optima.

\paragraph{Optimization Dynamics.}
To elucidate the ability of OptVQ to resolve local pitfalls in VQNs, we depict the dynamic training trajectory in a two-dimensional context, as illustrated in \cref{fig:exp_process}.
We initiate $100$ data points and $25$ codes at random, where the vanilla VQ with the nearest neighbor method is denoted by red and OptVQ with the optimal transport method is denoted by green.
It is evident that vanilla VQ is markedly influenced by the initial distribution, engaging predominantly with codes located at the distribition's extremities.
In contrast, OptVQ effectively harnesses the global structure, resulting in an equitable matching between data points and codes.

\paragraph{Consistency Check.}
We conduct an analysis to assess the consistency between OptVQ and the vanilla VQ in \cref{fig:exp_consistency}.
When there is a substantial divergence between the distribution of data points and codes, OptVQ exhibits enhanced global properties, ensuring that the matching between data pionts and codes is guided by global structure, thus circumventing the local pitfalls.
Conversely, in cases where the distribution of data points closely mirrors that of codes, OptVQ's performance is congruent with the nearest neighbor method, which is consistent with out expectations.

\subsection{Ablation Study} \label{sec:exp_abl}

In this section, we conduct a series of ablation studies to ascertain the impact of optimal transport operations on codebook utilization and reconstruction performance, as well as to evaluate the impact of normalization and the number of iterations on the Sinkhorn algorithm.

\begin{figure}[tbp]
    \centering
    \subfloat[Without normalization\label{fig:abl_not_norm}]{
        \includegraphics[width=0.42\linewidth]{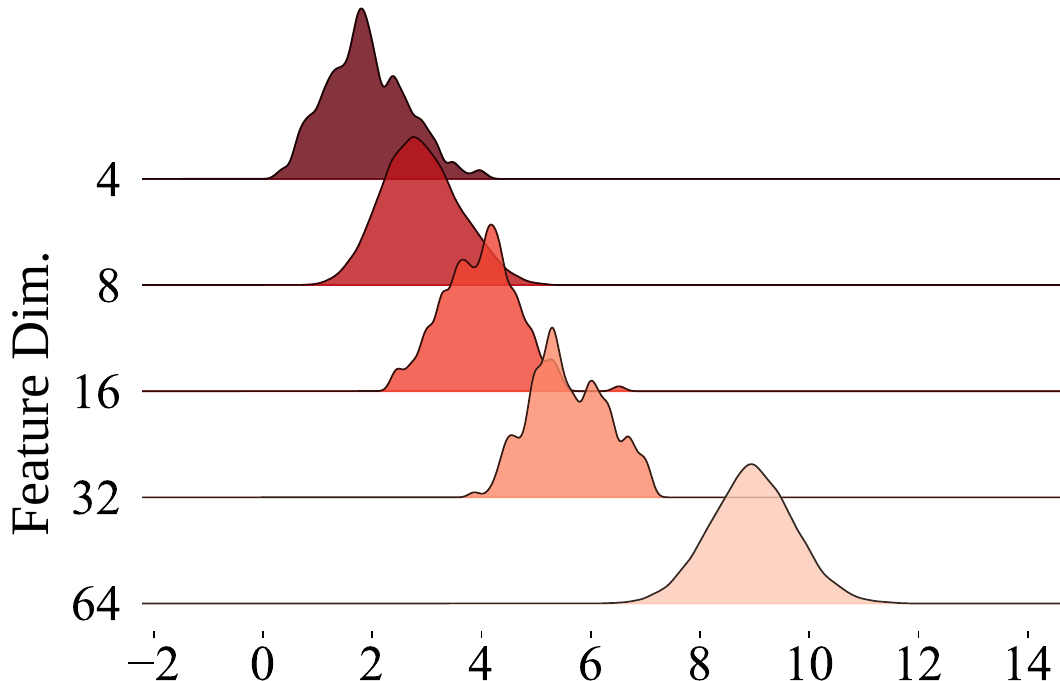}
    }
    \subfloat[With normalization\label{fig:abl_norm}]{
        \includegraphics[width=0.42\linewidth]{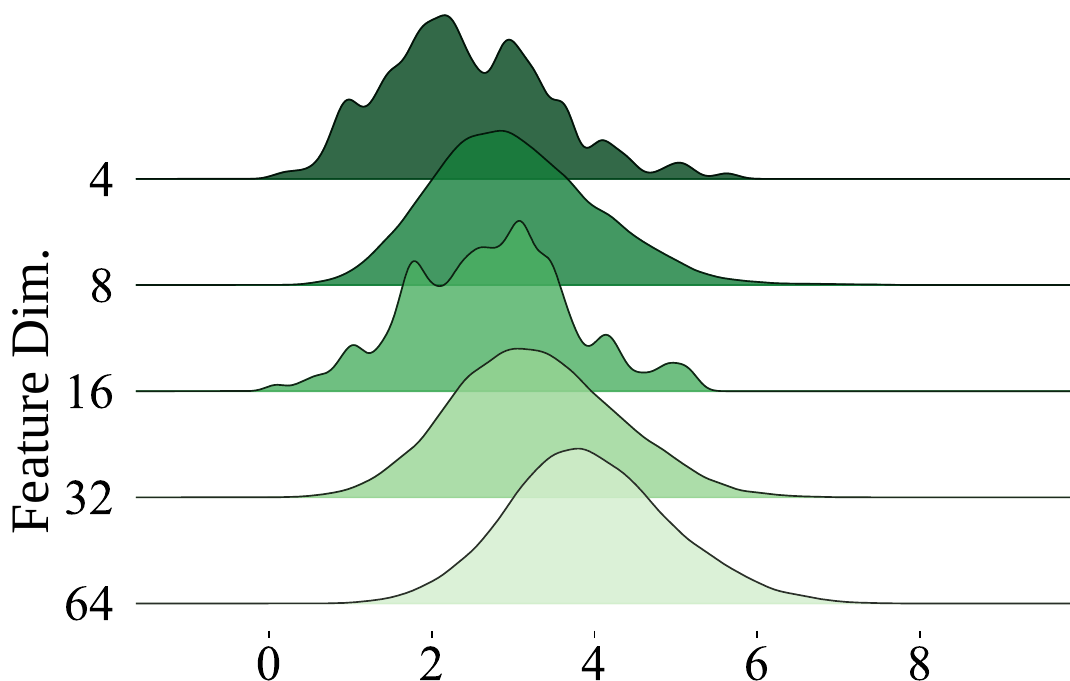}
    }
    \vspace{-2mm}
    \caption{Impact of normalization.}
    \vspace{-4mm}
    \label{fig:exp_norm}
\end{figure}

\vspace{-3mm}
\paragraph{Optimal Transport v.s. Nearest Neighbor.}
We initiate our analysis by comparing the optimal transport operation with the nearest neighbor across varying conditions in \cref{tab:exp_abl_sinkhorn}, including different latent dimensions and codebook sizes.
These experiments are conducted on a 1\% subset of ImageNet for efficiency, with all other parameters aligning with those previously described.
The conventional VQ-VAE experiences a substantial decrease in codebook utilization as the codebook size increases, plummeting below 1\%.
In contrast, OptVQ consistently maintains a 100\% codebook utilization, leading to a marked improvement in reconstruction performance with an expanding codebook size.

\vspace{-3mm}
\paragraph{Normalization for Diverse Distributions.}
As discussed in \cref{sec:technique}, the Sinkhorn algorithm is susceptible to numerical instability.
We illustrate the impact of normalization with diverse distributions, as depicted in \cref{fig:exp_norm}.
We generate random distributions for various latent variable dimensions and compute the histograms for the distance matrix $\mD$.
It is evident that normalization effectively confine the range of distance values to a reasonable interval, thereby bolstering the numerical stability of the Sinkhorn algorithm.

\begin{figure}[tbp]
    \centering
    \includegraphics[width=\linewidth]{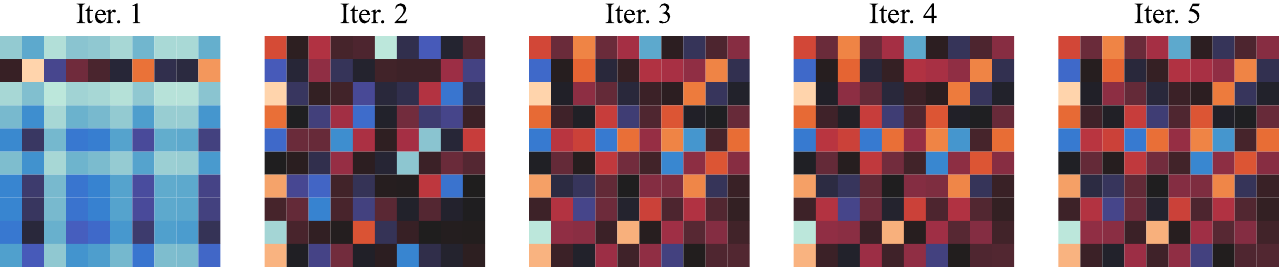}
    \caption{Convergence analysis for Sinkhorn iterations.}
    \vspace{-3mm}
    \label{fig:exp_iters}
\end{figure}

\paragraph{Iterations for Sinkhorn Algorithm.}
We probe the convergence of the Sinkhorn algorithm in \cref{fig:exp_iters}, adhering to the same settings with $\epsilon=10$.
We randomly simulated $10$ data points and $10$ codes with a dimension of $16$, and then execute the Sinkhorn algorithm to scrutinize the $\mA^t$ matrix post each iteration.
Our observations reveal that the values in $\mA^t$ progressively converge as the number of iterations increases.
Notably, the Sinkhorn algorithm has achieved convergence by the fifth iteration.
Thus, a relatively small number of iterations suffices to yield a reasonably accurate solution for the optimal transport problem.

\section{Discussion}

In this study, we tackle the local optimization obstables in conventional VQNs.
To this end, we introduce the OptVQ, which leverages global structure.
This method conceptualizes vector quantization as an optimal matching problem between data points and codes.
It efficiently attains optimal matching via the Sinkhorn algorithm, requiring only a minimal number of iterations.
Furthermore, we have conduct visualization experiments to scrutinize the behavior of OptVQ during the optimaization processs, thereby substatiating its effectiveness in overcoming local pitfalls.
Our quantitative experiments on ImageNet, MNIST, and CIFAR-10 demonstrate that OptVQ achieves a full utilization of codebook, which leads to superior reconstruction performance.

We also recognize that OptVQ is not without limitations.
Certain parameters in OptVQ, such as $\epsilon$, remain dependent on manual tuning despite being decoupled from input distributions.
We aim to delve into the theoretical exploration of the parameter choices.
Owing to limitations in computational resources, the scope of this paper is confined to the reconstruction performance.
We have not yet expanded the application of OptVQ to encompass more extensive visual tasks, such as image generation.
We are committed to investigating the potential of OptVQ in broader-scale applications in our forthcoming research.

{
    \small
    \bibliographystyle{ieeenat_fullname}
    \bibliography{main}
}

\appendix
\clearpage

\onecolumn
\startcontents[sections]
\printcontents[sections]{l}{1}{
    \setcounter{tocdepth}{2}
    \section*{\centering Table of Content for Appendix \rule{\linewidth}{0.5pt}}
}
\rule{\linewidth}{0.5pt}

\section{Algorithm Details}

\begin{algorithm}[htbp]
    \caption{The OptVQ algorithm.}
    \label{alg:optvq}
    \begin{algorithmic}[1]
    \renewcommand{\algorithmicrequire}{\textbf{Input:}}
    \renewcommand{\algorithmicensure}{\textbf{Output:}}
    \REQUIRE Continuous feature vector $\vz$, the codebook $\mathcal{C} = \{\vc_1, \ldots, \vc_n\}$, coefficient $\epsilon$, iteration number $T$.
    \ENSURE  Quantized feature vector $\vz_q$.
    \\ \textit{Initialization}:
       \STATE Compute the distance matrix $\mD$ between $\vz$ and $\mathcal{C}$.
       \STATE Normalize the matrix $\mD''$ as \cref{equ:normalization_1,equ:normalization_2}.
       \STATE Initiate the assignment matrix $\mA^0 = e^{- \epsilon \mD''}$ and $I=1$.
    \\ \textit{LOOP Process}:
       \FOR {$I \leq T$}
          \STATE Normalize the row: $\mA^{t+1}_{ij} = \mA^t_{ij} / \sum_k \mA^t_{ik}$.
          \STATE Normalize the column: $\mA^{t+2}_{ij} = \mA^{t+1}_{ij} / \sum_k \mA^{t+1}_{kj}$.
          \STATE Update index $I \leftarrow I + 1$.
       \ENDFOR
       \STATE Get the final quantized vector $\vz_q$ as \cref{equ:optvq}.
    \RETURN $\vz_q$
    \end{algorithmic}
\end{algorithm}

In this section, we delineate the details of the OptVQ algorithm, as depicted in \cref{alg:optvq}.
Initially, the OptVQ algorithm computes the distance matrix $\mD$ and subsequently normalizes it using \cref{equ:normalization_1,equ:normalization_2} to obtain the normalized matrix $\mD''$.
Prior to the commencement of the iterative process, the assignment matrix $\mA^0$ is initialized.
Thereafter, the algorithm iteratively updates the assignment matrix $\mA^t$ through an alternating normalization of rows and columns, culminating in the quantized feature vector $\vz_q$.

\section{Model Structure}

\begin{table}[htbp]
    \centering
    \caption{Model structure for MNIST and CIFAR-10.}
    \begin{tabular}{|c|l|l|}
        \hline
        \multirow{2}[4]{*}{\textbf{Model}} 
        & \multicolumn{1}{c|}{\multirow{2}[4]{*}{\textbf{Layer}}} 
        & \textbf{for MNIST and CIFAR-10} \bigstrut\\
        \cline{3-3} & & \boldmath{}\textbf{Input 32 $\times$ 32 image}\unboldmath{} \bigstrut\\
        \hline
        \multirow{6}[2]{*}{\textbf{Encoder}} & conv1 & 3 $\times$ 3, 16, padding 1 + ReLU \bigstrut[t]\\
            & down1 & Conv: 2 $\times$ 2, stride 2 \\
            & conv2 & 3 $\times$ 3, 16, padding 1 + ReLU \\
            & down2 & Conv: 2 $\times$ 2, stride 2 \\
            & conv3 & 3 $\times$ 3, 32, padding 1 + ReLU \\
            & conv4 & 3 $\times$ 3, 32, padding 1 \bigstrut[b]\\
        \hline
        \multirow{6}[2]{*}{\textbf{Decoder}} & conv1 & 3 $\times$ 3, 32, padding 1 + ReLU \bigstrut[t]\\
            & up1   & ConvT: 2 $\times$ 2, 16, stride 2 \\
            & conv2 & 3 $\times$ 3, 16, padding 1 + ReLU \\
            & up2   & ConvT: 2 $\times$ 2, 16, stride 2 \\
            & conv3 & 3 $\times$ 3, 16, padding 1 + ReLU \\
            & conv4 & 3 $\times$ 3, 1/3, padding 1 + ReLU \bigstrut[b]\\
        \hline
    \end{tabular}
    \label{tab:app_model_structure}
\end{table}%

In this paper, we employ two model architectures.
For the ImageNet dataset, we have adhered to the model structure proposed in VQGAN~\cite{esser2021taming}.
For in-depth insights into this particular model, we direct the readers to the seminal work.
Concurrently, for the MNIST and CIFAR-10 datasets, we have crafted a straightforward encoder-decoder configuration characterized by a downsampling factor denoted as $f=4$, the specifics of which are delineated in \cref{tab:app_model_structure}.
Within this model, the downsampling states are facilitated by convolutional layers that incorporate a kernel size and stride both set to $2$.
Conversely, the upsampling states are executed through the complementary transposed convolutional layers.

\section{Codebook Utilization}

\begin{figure*}[htbp]
    \centering
    \includegraphics[width=\textwidth]{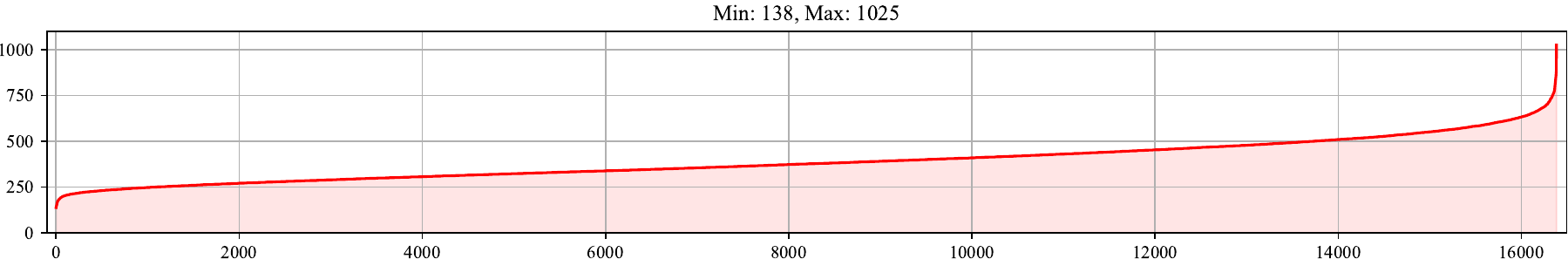}
    \caption{Codebook utilization}
    \label{fig:app_codebook_utilization}
\end{figure*}

In this paper, we have presented statistical data pertaining to the overall codebook utilization rate.
Building upon this, we now offer a granular analysis of the codebook utilization statistics, as illustrated in \cref{fig:app_codebook_utilization}.
To elaborate, for the OptVQ model that has been trained, we undertook a comprehensive analysis to ascertain the frequency of selection for each code in the codebook during the quantization process on the ImageNet validation set, which consists of 50,000 images.
Our findings reveal that all codebooks are engaged, exhibiting a relatively consistent pattern of utilization.
The least frequently utilized code is selected 138 times, while the most frequently utilized is chosen 1,025 times.
Notably, the majority of codes are selected between 300 to 600 times.
These observations underscore the OptVQ method's ability to fully harness the capacity of the codebook, resulting in a more balanced distribution of codebook usage.
Consequently, this effectively addresses the local issue that is commonly encountered in conventional VQNs.

\section{Training Statistics}

\begin{figure}[htbp]
    \centering
    \subfloat[Reconstruction Loss\label{fig:app_train_rec_loss}]{
        \includegraphics[width=0.33\linewidth]{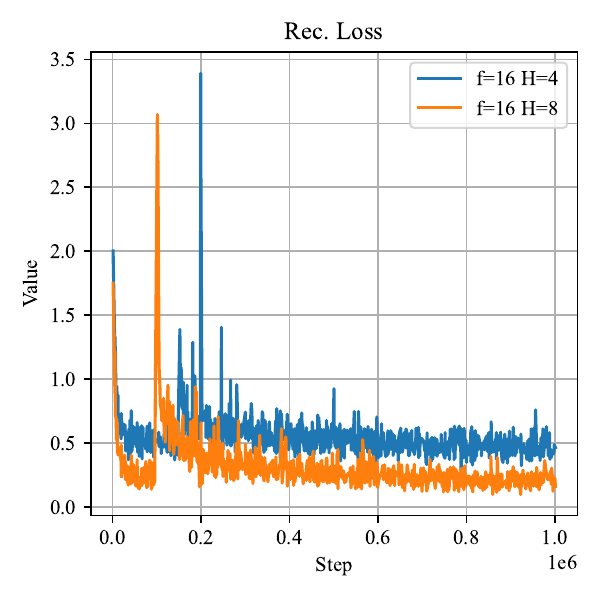}
    }
    \subfloat[Codebook Utilization\label{fig:app_train_codebook_usage}]{
        \includegraphics[width=0.33\linewidth]{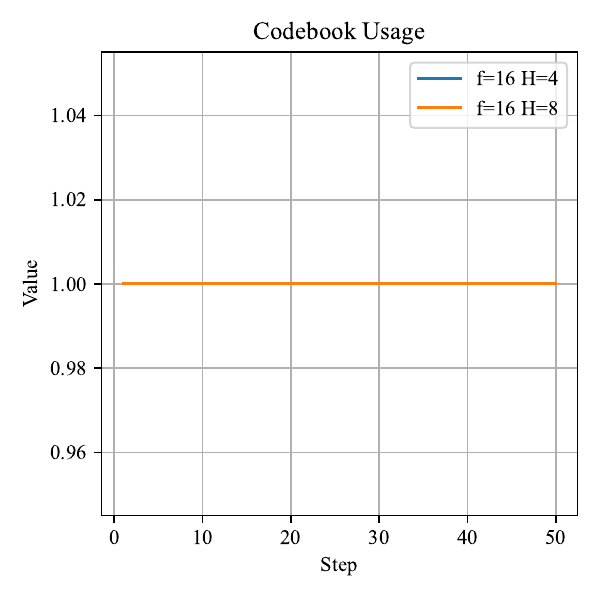}
    }
    \subfloat[L1 Loss\label{fig:app_train_l1}]{
        \includegraphics[width=0.33\linewidth]{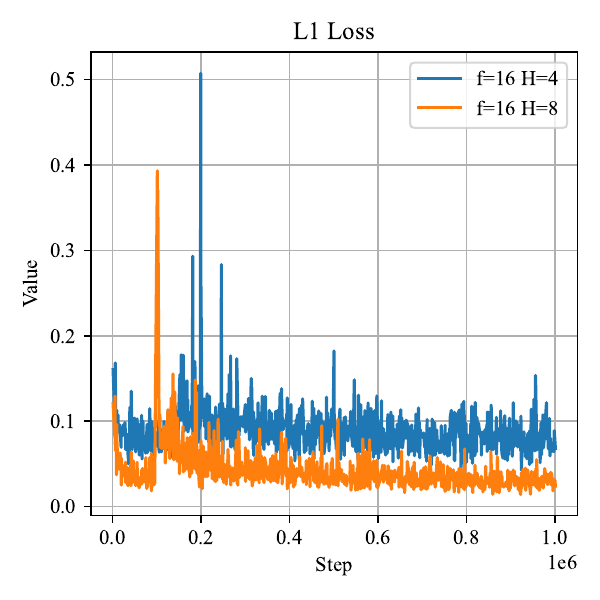}
    }
    //
    \subfloat[L2 Loss\label{fig:app_train_l2}]{
        \includegraphics[width=0.33\linewidth]{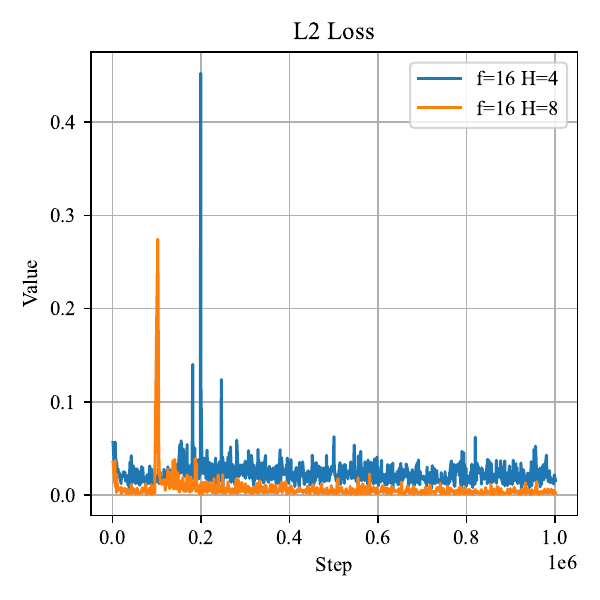}
    }
    \subfloat[Perceptual Loss\label{fig:app_train_p}]{
        \includegraphics[width=0.33\linewidth]{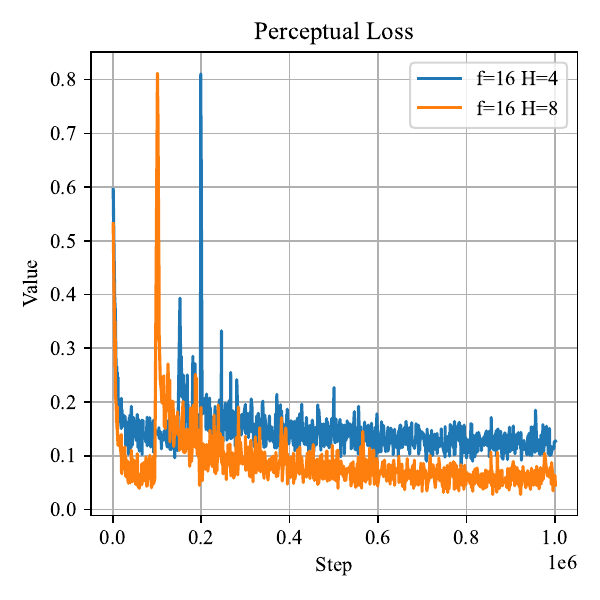}
    }
    \subfloat[Codebook Loss\label{fig:app_train_codeloss}]{
        \includegraphics[width=0.33\linewidth]{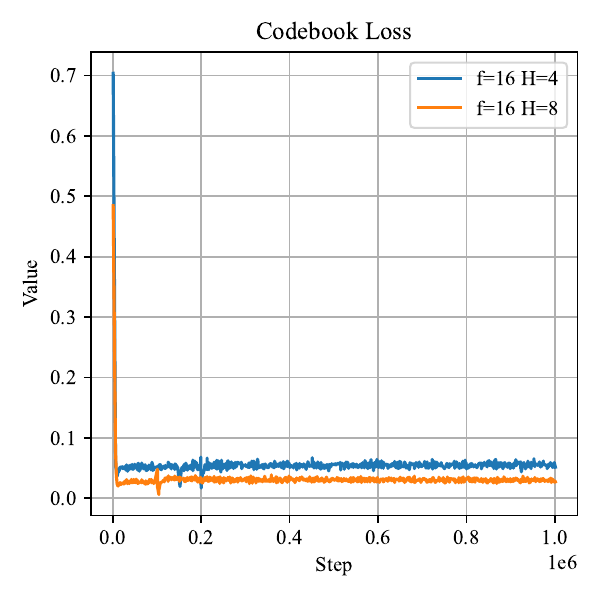}
    }
    //
    \subfloat[GAN Loss\label{fig:app_train_gan}]{
        \includegraphics[width=0.33\linewidth]{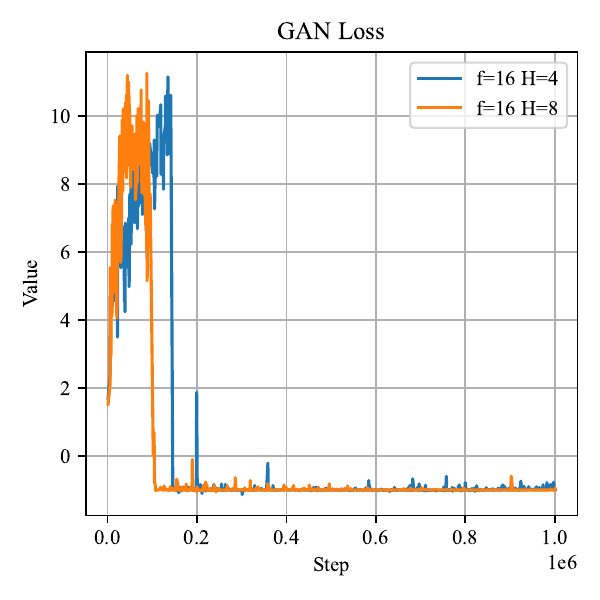}
    }
    \subfloat[Discriminator Loss\label{fig:app_train_disc}]{
        \includegraphics[width=0.33\linewidth]{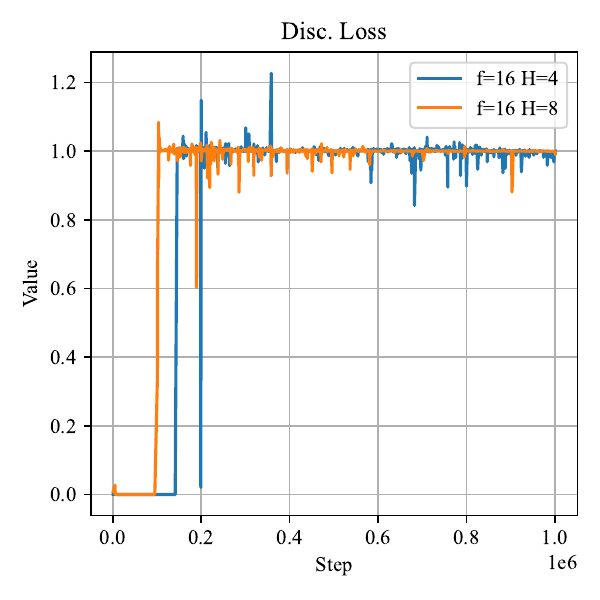}
    }
    \subfloat[GAN Loss Weight\label{fig:app_train_gan_weight}]{
        \includegraphics[width=0.33\linewidth]{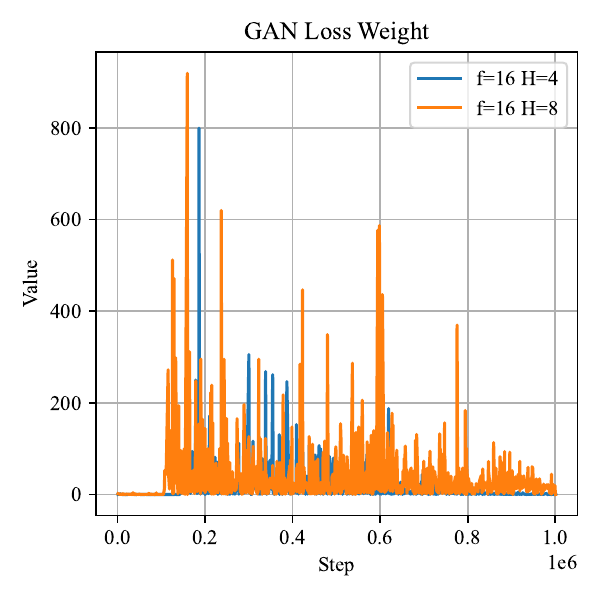}
    }
    \caption{Details training statistics.}
    \label{fig:app_train_statistics}
\end{figure}

In this section, we present a comprehensive set of training statistics to facilitate a deeper understanding of our experiments.
\cref{fig:app_train_rec_loss} delineates the trajectory of the reconstruction loss, which is a composite of L1 loss, L2 loss, and perceptual loss, each contributing to the weighted sum. 
The observed fluctuations within the curve are indicative of the onset of GAN training dynamics.
\cref{fig:app_train_l1}, \cref{fig:app_train_l2}, and \cref{fig:app_train_p} respectively chart the evolution of L1 loss, L2 loss, and perceptual loss throughout the training. 
It is evident that the perceptual loss experiences the most pronounced reduction, succeeded by L1 loss, with L2 loss exhibiting the least variation.
\cref{fig:app_train_codebook_usage} attests to the consistent 100\% utilization of the codebook throughout the training, signifying the optimal exploitation of the codebook's capacity. 
The behavior of the codebook loss is captured in \cref{fig:app_train_codeloss}, which indicates a swift descent and rapid convergence at the initial phase of training.
Furthermore, \cref{fig:app_train_gan} and \cref{fig:app_train_disc} respectively exhibit the progression of GAN loss and discriminator loss. 
It is observed that, as the discriminator is constituted by a rudimentary convolutional network, it becomes progressively outmaneuvered by the generator, culminating in a stabilization of the discriminator's loss value in the advanced stages of training.

\section{More Reconstruction Results}

\begin{figure}[htbp]
    \centering
    \includegraphics[width=0.88\linewidth]{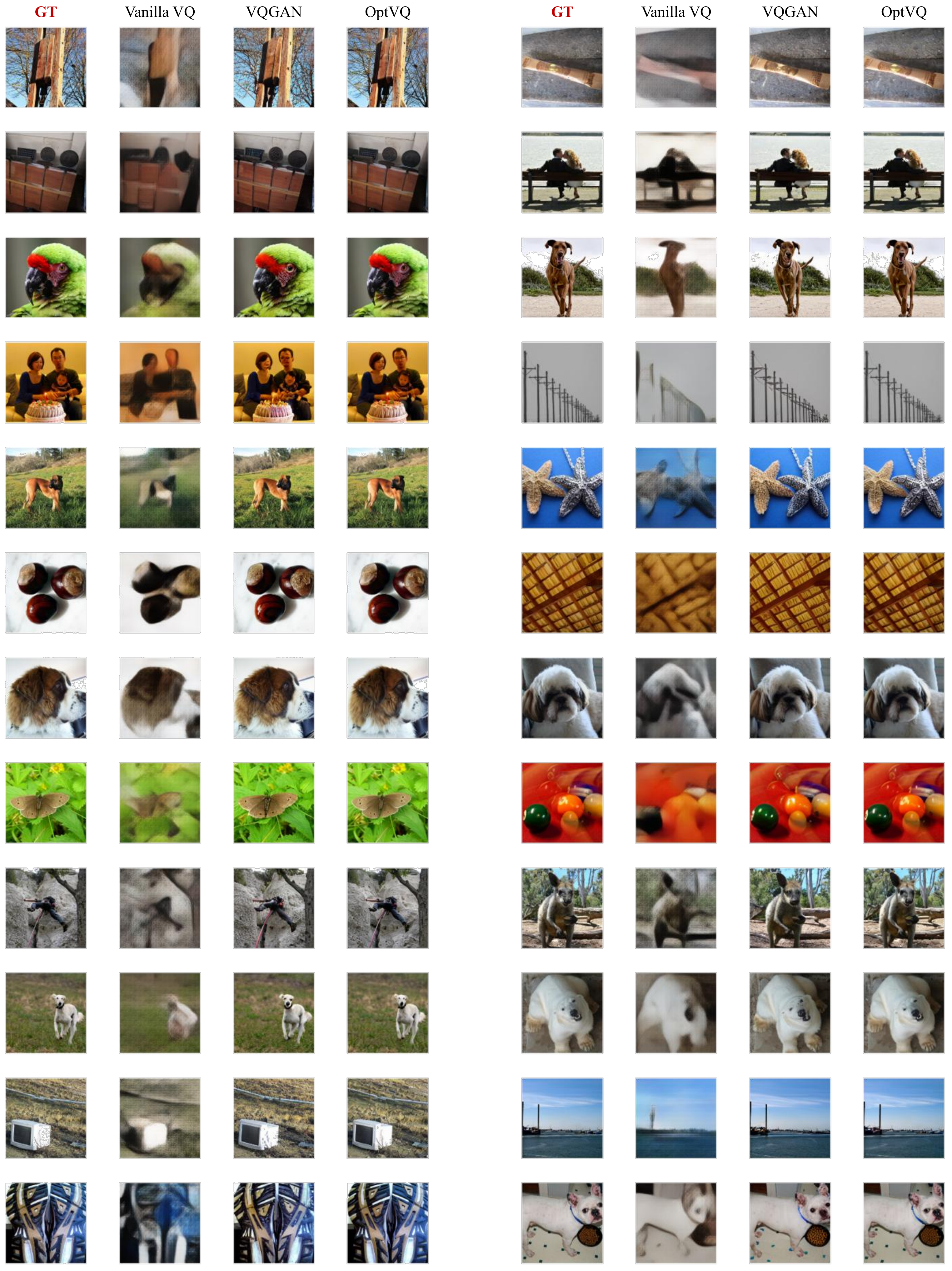}
    \caption{Additional reconstruction results.}
    \label{fig:app_reconstruction}
\end{figure}

In this section, we will provide more reconstruction results, as shown in \cref{fig:app_reconstruction}.
These results further prove the superior performance of OptVQ model on ImageNet dataset.

\end{document}